\documentclass[letterpaper, 10 pt, conference]{ieeeconf}  % Comment this line out if you need a4paper
\IEEEoverridecommandlockouts
\usepackage{bbm}
\usepackage[utf8]{inputenc} % allow utf-8 input
\usepackage[T1]{fontenc}    % use 8-bit T1 fonts
\usepackage{amsmath,amssymb,amsfonts}
\usepackage{hyperref}       % hyperlinks
\usepackage{url}            % simple URL typesetting
\usepackage{booktabs}       % professional-quality tables
\usepackage{nicefrac}       % compact symbols for 1/2, etc.
\usepackage{microtype}      % microtypography
\usepackage{xcolor}         % colors
\usepackage{algorithm}
\usepackage{algpseudocode}
\usepackage{graphicx}
\usepackage{subcaption}
\usepackage{tabularx}
\usepackage{xcolor} % Include the xcolor package

\usepackage{placeins}
\usepackage{amsthm}
\usepackage[backend=bibtex,style=numeric,natbib=true,style=ieee,maxcitenames=1,mincitenames=1]{biblatex}
\theoremstyle{definition}
\theoremstyle{plain}
\newtheorem{theorem}{Theorem}

\theoremstyle{definition}
\newtheorem{definition}[theorem]{Definition}
\theoremstyle{remark}
\newtheorem{remark}[theorem]{Remark}

\definecolor{red}{RGB}{255, 0, 0}
\algnewcommand{\LeftComment}[1]{\State \(\triangleright\) #1}
\usepackage{caption}

\begin{document}
\title{Neural Lyapunov Function Approximation with Self-Supervised Reinforcement Learning}

\author{Luc McCutcheon$^1$, Bahman Gharesifard$^2$, Saber Fallah$^1$
% \author{\IEEEauthorblockN{1\textsuperscript{st} Luc McCutcheon}
% \IEEEauthorblockA{\textit{Department of Mechanical Engineering Sciences} \\
% \textit{University of Surrey}\\
% \and
% email address or ORCID}}
\thanks{$^{1}$Luc McCutcheon and Saber Fallah are with CAV-Lab, Department of Mechanical Engineering Sciences, University of Surrey,\texttt{\{ lm01065, s.fallah \}@surrey.ac.uk}, 
% $^{2}$Liqun Zaho, University of Oxford, 
Bahman Gharesifard, Department of Electrical and Computer Engineering, University of California, Los Angeles \texttt{gharesifard@ucla.edu}
}
}
\maketitle
% ----------------------------------------------------------------------------------------------------------------------------------------------------

\begin{abstract}
Control Lyapunov functions are traditionally used to design a controller which ensures convergence to a desired state, yet deriving these functions for nonlinear systems remains a complex challenge. 
This paper presents a novel, sample-efficient method for neural approximation of nonlinear Lyapunov functions, leveraging self-supervised Reinforcement Learning (RL) to enhance training data generation, particularly for inaccurately represented regions of the state space. The proposed approach employs a data-driven World Model to train Lyapunov functions from \emph{off-policy} trajectories. The method is validated on both standard and goal-conditioned robotic tasks, demonstrating faster convergence and higher approximation accuracy compared to the state-of-the-art neural Lyapunov approximation baseline. 
The code is available at: \url{https://github.com/CAV-Research-Lab/SACLA.git}
\end{abstract}

\section{Introduction}

Safety-critical applications such as robotic surgery, autonomous vehicles, and hazardous material handling require controllers that provide stability and robustness under uncertainties. For nonlinear systems, this typically involves constructing a Lyapunov function. This scalar function decreases along system trajectories and certifies stability around an equilibrium point. However, identifying valid Lyapunov functions for nonlinear systems is challenging, often requiring significant expertise and compute \cite{lavaei2023systematic}. Traditional methods, like manual examination or Sum-of-Squares programs \cite{anderson2015advances}, are inefficient and do not scale to high-dimensional systems.

Neural Lyapunov Functions (NLFs) have been introduced \cite{chang2019neural, yang2024lyapunov, discreteNLC2023}, leveraging neural networks to approximate Lyapunov functions through iterative sampling. Despite their potential, existing methods for validating NLFs \cite{gao2012delta} become computationally intractable as state spaces expand.
While prior research primarily focuses on leveraging NLFs to stabilize the controller learning process, we explore an alternative perspective: \emph{How can the controller learning process be leveraged to facilitate the learning of NLFs?} In this work Reinforcement Learning (RL) is used to enhance the accuracy and applicability of NLFs in complex systems.

A novel method for online NLF approximation is proposed for discrete-time systems formulated as goal-conditioned Markov Decision Processes (MDPs). This extension allows for flexible equilibrium state configuration. Secondly, the proposed framework consists of an \emph{off-policy} RL agent that, given a \emph{self-supervised} objective, is incentivized to explore regions of instability throughout training, which is shown to increase the \textit{Region of Attraction (ROA)} and robustness of the controller at test-time. We define the ROA as the set of states around the goal location from which the system exhibits a decreasing Lyapunov value along trajectories, indicating movement toward the goal state.

% Unlike previous methods \cite{AlmostLyapCritics, zhou2022neural} that restrict exploration to stable regions during training, we focus on test-time stability and demonstrate that NLF-informed exploration can enhance the stability of the resultant controller.
Finally, to increase the scalability, we forgo strong certification in favor of a general-purpose stability metric. Since the Lyapunov function is intrinsically linked to the controlled system dynamics, a World Model (WM) is used to simulate on-policy Lie derivatives, from off-policy interactions, reducing the variance in Lie derivative estimates. Finally, this study demonstrate how the WM can be used to analyze the stability of the resultant system.

In summary, the primary contributions of this paper are:

\begin{itemize} \item \textbf{Self-Supervised RL Objective:} We introduce a self-supervised RL objective that incentivizes exploration into regions of instability throughout training, which increases the ROA and robustness of the NLF at test-time.
\item \textbf{Extension of Existing Frameworks:} We extend the Almost Lyapunov Critic framework \cite{AlmostLyapCritics} to an off-policy, goal-conditioned setting using a learned probabilistic model of system dynamics, improving sample efficiency and accommodating varying equilibrium points. 
\item \textbf{Comprehensive Stability Analysis:} We conduct an in-depth stability analysis of our proposed methods, demonstrating their effectiveness in assessing the stability of nonlinear control systems without necessitating environment interactions. 
\end{itemize}

% ----------------------------------------------------------------------------------------------------------------------------------------------------
\section{Related work} %% RELATED WORK ------------------------------------------------------------------------------------------------------------------

%% NLC
NLFs have emerged as a powerful technique for approximating Lyapunov functions in complex systems where derivation or search methods are intractable \cite{lavaei2023systematic}. By leveraging neural networks, NLFs capture intricate nonlinear dynamics, facilitating stability analysis and controller design. The concept was formally introduced by \citet{1994lyapunovMFirst} and popularized by \citet{chang2019neural}, who proposed the notion of Lyapunov risk for training NLFs. Subsequent works, such as \citet{discreteNLC2023}, extended NLFs to discrete-time systems, enhancing their applicability. However, these methods often rely on computationally intensive verification procedures. Additionally, some approaches assume restrictive forms for Lyapunov functions, such as quadratic structures \cite{chen2021learning}, which may not suit highly nonlinear systems \cite{yang2024lyapunov}.
 
%% Lyap Critic
Integrating Lyapunov functions with RL has shown promise in enhancing the stability of learned controllers. Studies by \citet{AlmostLyapCritics} and \citet{hanActorCriticReinforcementLearning2020} explored learning Lyapunov functions alongside control policies to improve stability throughout training, though results demonstrate NLF learning accuracy is an active challenge. Furthermore, NLFs have been utilized to enforce safety in Constrained Markov Decision Processes, reducing constraint violations \cite{altman2021constrained, chow2018lyapunov}.

%WM
While Lyapunov-based stability methods have been explored in RL, most approaches assume system dynamics are partially known \emph{a priori}, which limits general applicability \cite{berkenkamp2017safe, gros2020safe}. This assumption of a known dynamics prior, however, cannot always be provided and introduces additional bias. Therefore, data-driven WMs have been used to enhance RL agents through modeling environmental uncertainties to improve decision-making and planning \cite{vlastelica2023mind}. WMs have been combined with Lyapunov functions to construct control barrier functions \cite{dawson2022safe, ames2019control, wang2023enforcing} and to facilitate safe exploration \cite{berkenkamp2017safe}; however they have not been used in the direct optimization of NLFs for test-time stability. These integrations enhance system safety and stability by providing robust future states predictions.

% Joint learning
Jointly learning control policies and dynamics models has been extensively explored to improve safety and efficiency in RL. Studies \cite{zhou2022neural,min2023data,zhang2024learning} have focused on simultaneously learning policies and system dynamics to enhance sample efficiency and ensure safe exploration. Additionally, \citet{kolter2019learning} investigated the joint learning of Lyapunov functions and dynamics without incorporating control policies. While these methods have advanced stable learning, they often face challenges related to computational expense and limited exploration. Our work however alleviates this computation cost through incrementally learning an NLF, using a probabilistic stability metric rather than relying on computationally intensive solvers for strong guarantees.

% Summary
Despite significant progress, existing methods for NLFs for RL face challenges in both scalability and robustness, especially in highly nonlinear systems. Moreover, many approaches impose restrictive assumptions on Lyapunov functions or depend on computationally intensive verification processes. By integrating a data-driven WM with RL, our approach improves efficiency, scalability, and robustness, addressing key limitations of existing methods while avoiding prohibitive computational costs.

 %This distinction sets our work apart from other Lyapunov-based methods that have been used in adversarial settings \cite{pinto2017robust, kang2021stable}, where the adversary directly perturbs the system through external forces to improve robustness.
% Add works on Riemannian Geometry

% ------------------------------------------------------------------------------------------------------------------------
\section{Preliminaries} % NOTATION ----------------------------------------------------------------------------------------------------
\label{Preliminaries}
This section introduces the fundamental concepts, notations, and definitions essential for understanding our proposed methodology.
\subsection{Markov Decision Process}
\label{sec:MDP}
  The proposed system considers an infinite-horizon discrete-time MDP, defined by the tuple ($\mathcal{X}$, $U$, $f$, $r$), with state space $\mathcal{X}$ and action space $U$ where $f(x_{t+1} \mid x_t,u_t)$ at time $t \in T$ denotes the stochastic transition dynamics from the current state $x_t \in \mathcal{X}$ to $x_{t+1} \in \mathcal{X}$ under a policy $\pi_\phi$ parameterized by $\phi$, from which actions $u_t$ are sampled $u_t\sim\pi_\phi(\cdot \mid x_t)$.
  
  Building upon the standard MDP framework, the notation is extended to accommodate \emph{goal-conditioned RL} \cite{liu2022gc_rlsurvey}. In this setting, a goal set $\mathcal{G} \subset \mathcal{X}$ is introduced, where each goal $g_t \in \mathcal{G}$ specifies a desired state that the agent aims to reach. Consequently, the reward function $r(x_t, g_t)$ and the policy $\pi_\phi(u_t \mid x_t, g_t)$ become dependent on the current goal $g_t$, allowing for more flexible and directed learning objectives.
  
The WM, denoted by $\tilde{f}_\xi$, is parameterized by $\xi$, and serves as an approximation of the true transition dynamics $f$. Specifically, the WM predicts the distribution of the next state $\tilde{x}_{t+1}$ given the current state $x_t$ and action $u_t$, i.e., $\tilde{f}_\xi(\tilde{x}_{t+1} \mid x_t, u_t) \approx f(x_{t+1} \mid x_t, u_t)$.

\subsection{Lyapunov Stability and Lie derivatives}

The Lie derivative $\textsf{L}_f V(x_t)$ measures the rate of change of a differentiable Lyapunov function $V: \mathcal{X} \rightarrow \mathbb{R}_{\geq 0}$ along the trajectories of the system dynamics $f$. In experiments, the Lie derivative are approximated using the WM as follows:
\begin{equation}
        \textsf{L}_{\tilde{f}}V(x_t) = \mathbb{E}_{\tilde{x}_{t+1}\sim \tilde{f}(\cdot\mid x_t,u_t)}[V(\tilde{x}_{t+1}) - V(x_t)]
\label{eq:Lie_derivative}
\end{equation}

Here, $\tilde{x}_{t+1}$ is the predicted next state generated by the WM. This approximation facilitates the computation of the Lie derivatives through sampled transitions from the replay buffer.

An investigation into the Lyapunov stability requires a formal definition of system stability around the goal point, which is given as:
% {\color{red}\bf B: Is this really correct? $ V$ should be positive, i.e., $ V(x)>0$ for all $ x\neq 0$, and you need to have a unique equilibrium ... }

\begin{definition}[$\epsilon$-stability]
    \label{def:stability}

    Consider a goal state $g_t \in \mathcal{G} \subset \mathcal{X}$. Define the region around $g_t$ as a ball $\mathcal{B}(g_t,\epsilon) = \{ x_t \in \mathcal{X} \mid ||x_t - g_t||_{\infty} < \epsilon \}$, where $\epsilon > 0$ denotes the radius of interest. A pair $(V, \pi_\phi)$, consisting of a Lyapunov function $V$ and a policy $\pi_\phi$, is said to be $\epsilon$-stable about $g_t$ if: (a) The Lie derivative satisfies $\textsf{L}_f V(x_t) < 0$ for all $x_t \in \mathcal{B}(g_t, \epsilon) \setminus \{g_t\}$ (b) $V(x_t) > 0$ for all $x_t \in \mathcal{B}(g_t, \epsilon) \setminus \{g_t\}$ (c) $V(g_t)=0$     
\end{definition}

% {\color{red} This condition ensures that the system's trajectories converge to the equilibrium point $g_t$ (see Equation \eqref{eq:Lie_derivative})}

% \end{definition}
% We define our NLF as an MLP parameterized by $\psi$. 
\begin{definition}[Lyapunov Risk] Consider the Lyapunov function, parameterized by $\psi$ for a controlled dynamic system. In order to quantify the extent to which the Lyapunov conditions are being violated, we define an objective function for the NLF based on the Lyapunov risk \cite{chang2019neural} as:

\begin{equation} 
    \label{eq:lyap_risk}
    J_V(\psi) = \mathbb{E}_{x_t,u_t \sim \pi_\phi(\cdot \mid x_t)} \bigg( -\bigg[ \max(0,\textsf{L}_{\tilde{f}}V_{\psi}(x_t)) + V^2_{\psi}(g_t) \bigg] \bigg)
\end{equation}

 The goal $g_t$ represents the equilibrium point. The $V^2_{\psi}(g_t)$ loss term encourages $V_{\psi}(g_t)=0$,  penalizing non-zero values at the goal state such that $g_t$ becomes the global minima. Secondly, the $\max(0,\textsf{L}_{\tilde{f}}V_{\psi}(x_t))$ term penalizes positive Lie derivatives. The positive definiteness of $V_\psi$ is enforced by designing the network architecture to output strictly positive values. This is achieved by taking the absolute value of the networks outputs and adding a small constant \cite{gaby2022lyapunov}. 
\end{definition}

Differentiating $J_V(\psi)$ with respect to the network parameters $\psi$ yields an unbiased estimator suitable for optimizing the NLF to better satisfy the Lyapunov conditions:

\begin{multline}
\label{eq:derivative_obj}
\nabla_{\psi} J_V(\psi) = \mathbb{E}_{x_t, u_t \sim \pi_\phi(\cdot \mid x_t)} \bigg( - \bigg[\\ \mathbf{1}(\textsf{L}_{\tilde{f}} V_{\psi}(x_t) > 0) \nabla_{\psi} (\textsf{L}_{\tilde{f}} V_{\psi}(x_t)) + 2 V_{\psi}(g_t) \nabla_{\psi} V_{\psi}(g_t) \bigg] \bigg)
\end{multline}

where $\mathbf{1}(\cdot)$ is the indicator function that equals 1 when $\textsf{L}_f V_\psi(x_t) > 0$ and 0 otherwise. This gradient facilitates the training of the NLF to minimize violations of the Lyapunov conditions.
\subsection{Probabilistic World Model}% PROBABALISTIC WM ----------------------------------------------------------------------------------------------------------------------
% See williams and Rasmussan 1996 and  https://proceedings.neurips.cc/paper_files/paper/2003/file/7993e11204b215b27694b6f139e34ce8-Paper.pdf
% !!Simplify like this paper: https://arxiv.org/pdf/2309.05582
WM's enable RL agents to anticipate and manage the inherent uncertainties of their environments. In this work, the WM learns a probability distribution over possible subsequent future states which we utilize for resampling off-policy transitions to calculate on-policy Lie derivatives (Equation ~\eqref{eq:Lie_derivative}). Thereby, reducing the variance in NLF training caused by the dynamics $f(x_{t+1} \mid x_t,u_t)$ depending on the current behavior policy $\pi_\phi$ and not the previous policies used to collect training data. The WM enhances the proposed approaches NLF approximation accuracy and provides valuable uncertainty estimates, which are crucial for robust decision-making under stochasticity, exemplified in Section \ref{sec:stab-analysis}. 

 The WM learns from transition data in the form of $\mathcal{D}=\{(x_t,u_t),x_{t+1}\}^T_{t=0}$ and is updated through minimizing the negative log likelihood loss function: 
\begin{equation}
     \mathrm{Loss}_\xi(x_t)=-\sum^T_{t=1}\log\tilde{f}_{\xi}(\Tilde{x}_{t+1} \mid x_t,u_t)
\end{equation}

The WM conditioned on $x_t$ and $u_t$, outputs a Gaussian distribution with a mean $\mu_\xi(x_t, u_t)$ and diagonal covariance matrix $\sigma_\xi(x_t, u_t)$. This implies that each element of $x_t$ is modeled independently and is uncorrelated with the rest of the state information. Such an assumption is commonly made in probabilistic models to simplify the modeling process \cite{chua2018PATS}.

\begin{equation}
    \tilde{f}_\xi(\tilde{x}_{t+1} \mid x_t, u_t) = \prod_{i=1}^n \mathcal{N}(\tilde{x}_{t+1,i}; \mu_{\xi,i}(x_t, u_t), \sigma^2_{\xi,i}(x_t, u_t))
    \label{eq:prob_dist_diag}
\end{equation}

Therefore, the loss is:
\begin{multline}
\label{eq:loss} % Not sure where to put the \\ here
\mathrm{Loss}_{\xi}(x_t) = \sum_{t=1}^T \left[ (\mu_{\xi}(x_t, u_t) - x_{t+1})^T \sigma^{-1}_{\xi}(x_t, u_t) \right. \\
\left. (\mu_{\xi}(x_t, u_t) - x_{t+1}) + \log \det \sigma_{\xi}(x_t, u_t) \right].
\end{multline}

\section{Policy Optimization and Lyapunov Approximation}
% Should be last section I write since it's dependant on results

In standard RL, the policy $\pi_\phi$ aims to maximize the expected $\gamma$-discounted cumulative return:
\begin{equation}
    \label{eq:classic-rl}
    J(\pi_\phi)=\mathbb{E}_{\pi_\phi, f}[\sum^\infty_{t=0}\gamma^tr(x_t,u_t)]
\end{equation}

This typical RL objective aims to maximize task-specific rewards, often without considering the system's stability. To promote system stability, we modify the policy objective by adding a regularization term that encourages exploration in regions where the Lyapunov function is poorly defined, thereby incentivizing exploration in unstable regions.

\begin{equation}
    \label{eq:mixed_obj}
    J(\pi_\phi) = \mathbb{E}_{\pi_\phi, f} \left[ \sum_{t=0}^{\infty} \left( (1 - \beta) \gamma^t r(x_t, g_t) + \beta \cdot \mathrm{Loss}_\psi(x_t) \right) \right]
\end{equation}

Here, $\beta \in [0,1]$ is a hyperparameter that balances the trade-off between reward and Lyapunov risk. In practice (See Figure \ref{fig:ip_grid}) we find $\beta=0.5$ to be work best. The Lyapunov loss, given as $\text{Loss}_\psi(x_t)$, is defined as:

\begin{equation}
    \mathrm{Loss}_{\psi}(x_t) = \max(0,\textsf{L}_{\tilde{f}}V_{\psi}(x_t)) + V^2_{\psi}(g_t) 
\end{equation}

This objective, is used by our Soft Actor-Critic Lyapunov Approximation (SACLA) method and incentivizes the agent to explore regions where the Lyapunov function is poorly defined, thereby enhancing the robustness and stability of the learned NLF.

We implement the SACLA algorithm, which leverages the Soft Actor-Critic (SAC) framework \cite{haarnoja2018soft} for online data generation and NLF learning. The algorithm integrates the following components:

\begin{itemize}
    \item \textbf{Policy Update:} Optimizes the policy $\pi_\phi$ to maximize the combined objective of reward and Lyapunov loss.
    \item \textbf{NLF Update:} Minimizes the Lyapunov risk $J_V(\psi)$ to improve the accuracy of the Lyapunov function.
    \item \textbf{WM Update:} Trains the WM $\tilde{f}_\xi$ to better predict state transitions, thereby reducing variance in Lie derivative estimates.
\end{itemize}

The detailed algorithm is presented in Algorithm \ref{alg:algorithm}.

\begin{algorithm}
\caption{Soft Actor-Critic Lyapunov Approximation}
\label{alg:algorithm}
\begin{algorithmic}
\LeftComment{Initial parameters}
\State \textbf{Input:} $\psi, \theta_1, \theta_2, \phi, \xi$ 
\begin{algorithmic}[1]
    
\State $\bar{\theta}_1 \leftarrow \theta_1, \bar{\theta}_2 \leftarrow \theta_2$ \LeftComment{Initialize target network weights}
\State $\mathcal{D} \leftarrow \emptyset$ \LeftComment{Initialize an empty replay pool}
\For{\texttt{each iteration}}
    \For{\texttt{each environment step}}
        \LeftComment{Sample action from the policy}
        \State $u_t \sim \pi_\phi(u_t \mid x_t)$ 
        \LeftComment{Sample transition from the environment}
        \State $x_{t+1} \sim f(x_{t+1} \mid x_t, u_t)$ 
        \LeftComment{Store the transition in the replay pool}
        \State $\mathcal{D} \leftarrow \mathcal{D} \cup \{(x_t, u_t, r(x_t, u_t), x_{t+1})\}$ 
    \EndFor
    \For{\texttt{each gradient step}}
    \LeftComment{Update NLF parameters using Equation \eqref{eq:derivative_obj}}
        \State $\psi \leftarrow \psi_V - \psi_V \hat{\nabla}_\psi J_V(\psi)$ 
        % \State $\xi \leftarrow \xi_f - \hat{\nabla}_\xi \arg\min_{\xi} \mathbb{E}_{\mathcal{D}}[-\log f_{\xi}(x_{t+1} |x_t, u_t)]$ \Comment{Update WM parameters}
        \LeftComment{Update WM parameters using Equation \eqref{eq:loss}}
        \State $\xi \leftarrow \xi - \lambda_\xi \nabla_\xi \mathrm{Loss}_{f_{\xi}}(\xi)$ 
        \LeftComment{Update the Q-function parameters}
        \State $\theta_i \leftarrow \theta_i - \lambda_Q \hat{\nabla}_{\theta_i} J_Q(\theta_i)$ for $i \in \{1, 2\}$ 
        \LeftComment{Update policy weights}
        \State $\phi \leftarrow \phi - \lambda_\pi \hat{\nabla}_\phi J_\pi(\phi)$ 
        \LeftComment{Adjust temperature}
        \State $\alpha \leftarrow \alpha - \lambda_\alpha \hat{\nabla}_\alpha J(\alpha)$ 
        \LeftComment{Update target network weights}
        \State $\bar{\theta}_i \leftarrow \tau \theta_i + (1 - \tau) \bar{\theta}_i$ for $i \in \{1, 2\}$ 
    \EndFor
\EndFor
\State \textbf{Output:} $\psi, \theta_1, \theta_2, \phi$
\end{algorithmic}
\end{algorithmic}
\end{algorithm}

\section{Experiments}
\label{sec:Experiments}
%For our experiments we utilize the classic control \texttt{InvertedPendulum-v4} environment with 
\subsection{Implementation Details}
All experiments have been conducted on a machine using an Intel i7-5820K CPU, a NVIDIA GeForce RTX 3080 GPU with Ubuntu 22.04 LTS. Methods have been validated using two key metrics: (1) The ROA is measured with the percentage of points within a defined region around the goal location that exhibits negative Lie derivatives, indicating stability. (2) The uncertainty introduced by the WM, assessed by the standard deviation of percentage negative Lie derivative predictions for each objective and task over intervals of 10 thousand time-steps. %For ROA calculations, we sampled 5,000 points uniformly distributed within a specified radius around the goal location to balance computational efficiency and statistical significance.

\subsection{Baselines}
We compared SACLA against the following baseline methods:
\begin{itemize}
    \item \textbf{Soft Actor-Critic (SAC):} Optimizes the expected discounted cumulative reward without incorporating Lyapunov constraints.
    \item \textbf{SACLA (\(\beta=1\)):} Guides the agent solely based on the NLF.
    \item \textbf{SACLA (\(\beta=0.5\)):} Balances reward maximization and NLF exploration, encouraging both task performance and stability.
    \item \textbf{Policy Optimization with Self-Learned Almost Lyapunov Critics (POLYC)} \cite{AlmostLyapCritics}: Integrates Lyapunov critics with policy optimization to enhance stability.
\end{itemize}
For POLYC, which is originally based on PPO \cite{schulman2017proximal}, the objective function was incorporated into our off-policy SACLA framework to ensure a fair comparison. POLYC's objective function penalizes positive Lie derivatives by including a term that rewards the agent when Lie derivatives are negative, effectively constraining exploration to regions that already satisfy stability conditions, which limits its ability to expand the ROA beyond initially stable regions.
    
% \begin{multline}    
%     \label{eq:polyc_eq}
%     J_{\text{P}}(\pi_\phi) = \mathbb{E}_{\pi_\phi, f} \bigg[ \sum_{t=0}^{\infty} \bigg( (1 - \beta)\gamma^t r(x_t, u_t) \\+ \beta \min(0, -\textsf{L}_f V(x_t)) \bigg) \bigg]
% \end{multline}

Throughout all experiments, the policy update step in Algorithm \ref{alg:algorithm} was replaced with the corresponding baseline policy updates.

%%%
\subsection{Region of Attraction}

To quantify the stability of each approach, we computed the ROA by evaluating the percentage of points within a defined grid around the goal location that have negative Lie derivatives. Specifically, for each environment:
\begin{itemize}
    \item \texttt{FetchReach-v2}: A grid spanning ±2 units along each Cartesian axis centered at the goal.
    \item \texttt{InvertedPendulum-v4}: A phase-space grid of angular displacement \(\theta\) and angular velocity \(\dot{\theta}\).
\end{itemize}

The percentage of negative Lie derivatives is calculated using:
\begin{equation}
\label{eq:percent_neg}
\% \text{ Negative } \textsf{L}_{\tilde{f}}V= \left( \frac{\sum_{i=1}^n \mathbf{1}(\textsf{L}_{\tilde{f}}V(x_i) < 0)}{n} \right) \times 100%
\end{equation}
where $\mathbf{1}(\cdot)$ is the indicator function. This metric aligns with Definition \ref{def:stability} and provides an estimate of the overall stability of the system in the vicinity of the goal.

\begin{figure}[htbp]
    \centering
    % First row of images
    \begin{subfigure}[b]{0.253\textwidth}
        \centering
        \includegraphics[width=\textwidth, trim=440 80 340 100, clip]{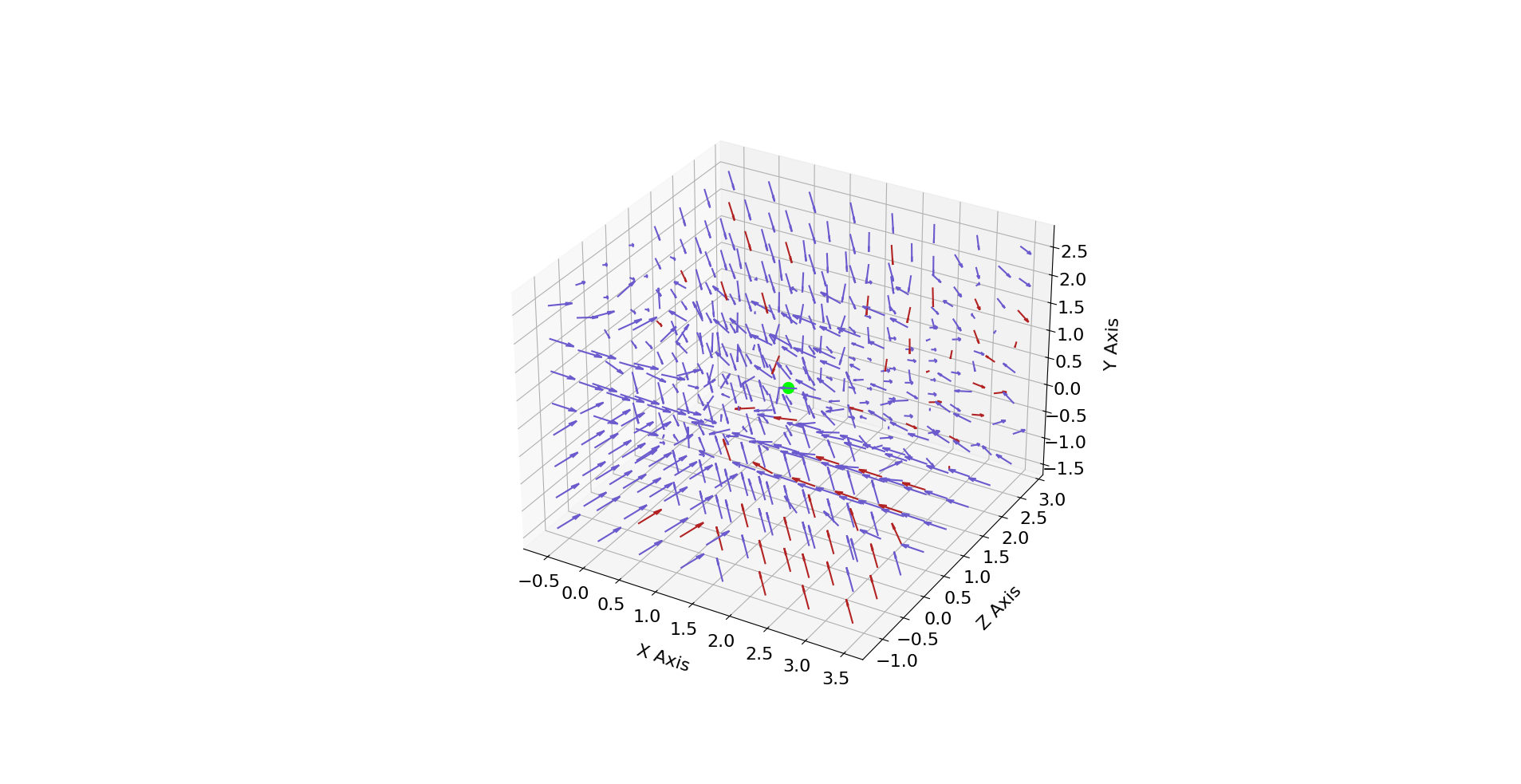}
        \captionsetup{justification=centering}
        \caption{ SACLA ($\beta=0.5$) ($88.87\% \pm 0.83$)}%$\pm$0.391\%)}
        \label{fig:fr_SACLA0.5}
    \end{subfigure}
    \hfill
    \begin{subfigure}[b]{0.226\textwidth}
        \centering
        \includegraphics[width=\textwidth, trim=440 80 300 100, clip]{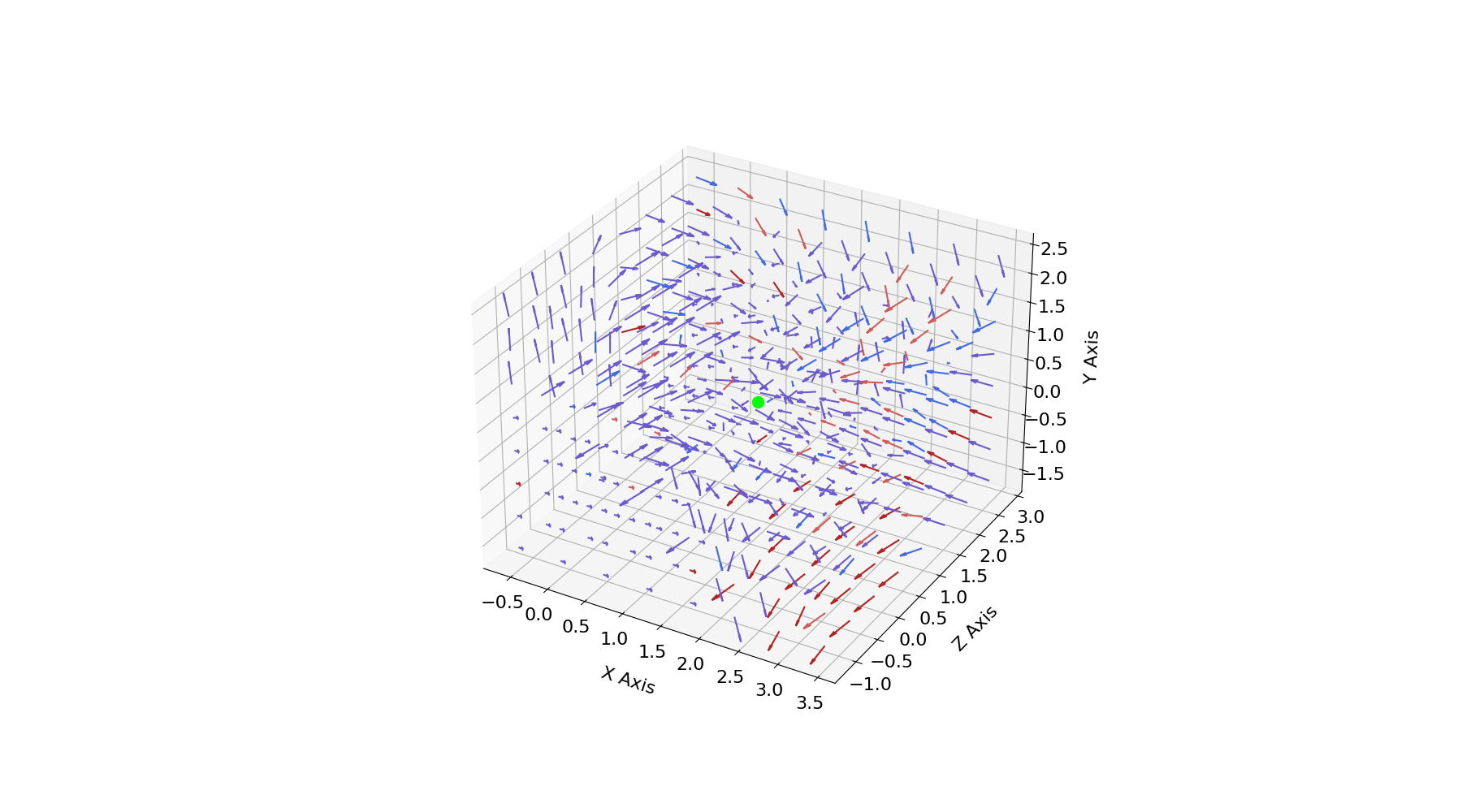}
        \captionsetup{justification=centering}
        \caption{SACLA ($\beta=1$)\\ ($85.55\% \pm 0.28$)}%$\pm$1.21\%)}
        \label{fig:fr_SACLA1}
    \end{subfigure}
    \hfill
    \begin{subfigure}[b]{0.253\textwidth}
        \centering
        \includegraphics[width=\textwidth, trim=440 80 340 100, clip]{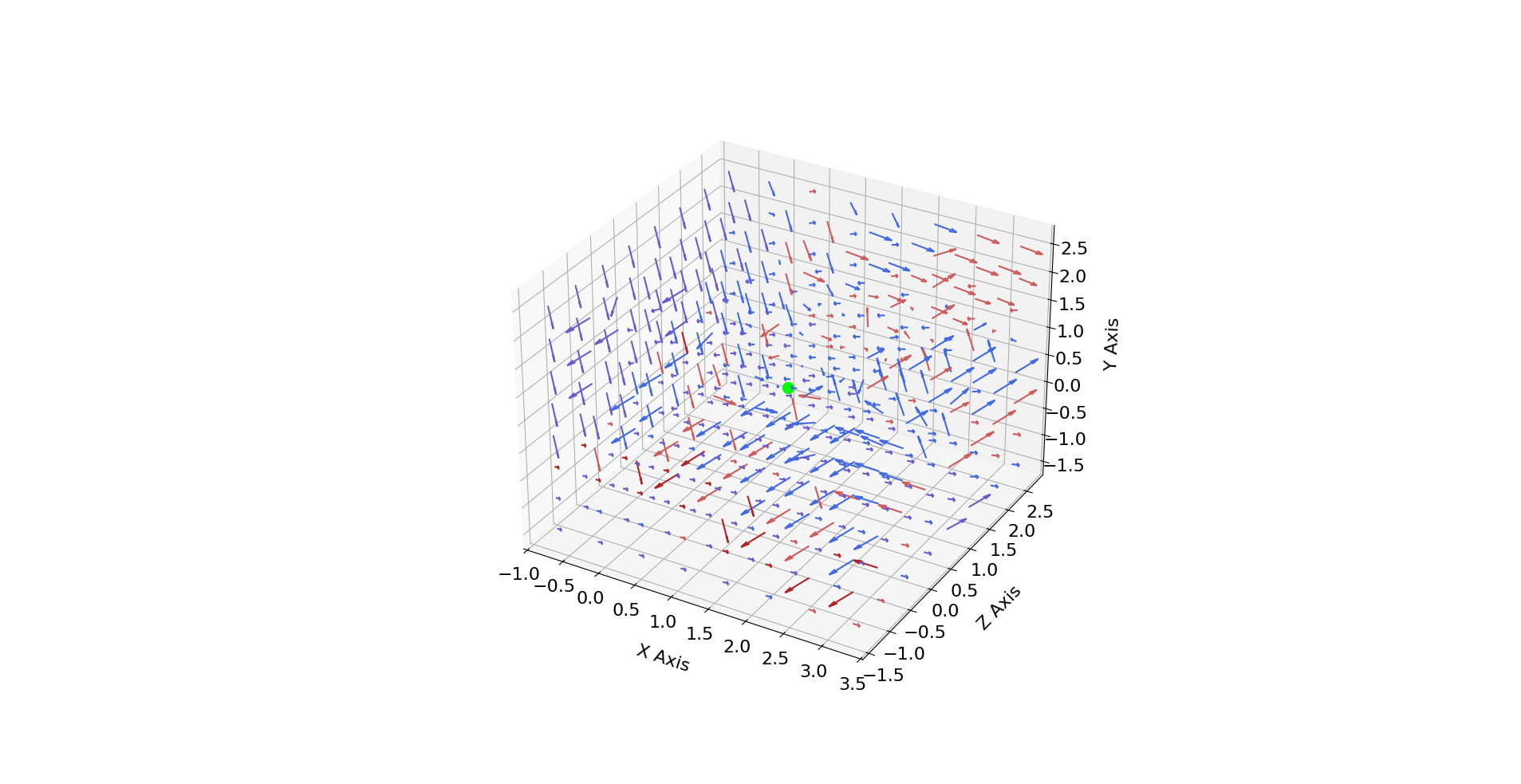}
        \caption{POLYC\\ $(76.75\% \pm 1.67)$}%\%)% $\pm$ 0.377\%)}
        \label{fig:fr_POLYC}
    \end{subfigure}
    \hfill
    \begin{subfigure}[b]{0.226\textwidth}
        \centering
        \includegraphics[width=\textwidth, trim=440 80 300 100, clip]{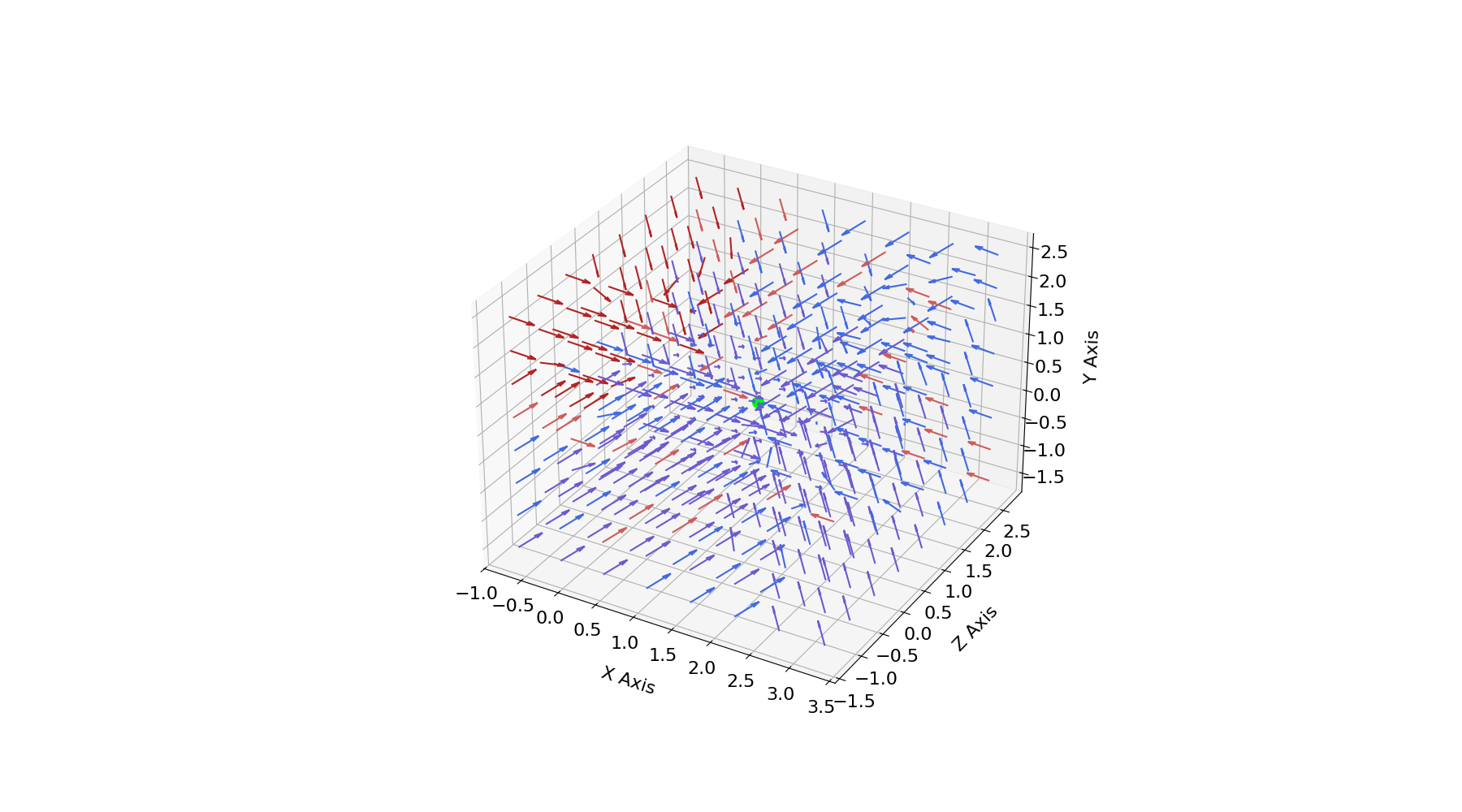}
        \caption{SAC\\ ($79.88\% \pm 1.10$)}%$\pm$0.431\%)}
        \label{fig:fr_SACL}
    \end{subfigure}
    \caption{ROA for SACLA in the \texttt{FetchReach-v2} environment for 512 points within 2 global coordinates of the goal location with the final performance percentage and standard deviation. Arrows indicate the action vector taken at each point. Blue arrows have negative Lie derivatives where as red have positive Lie derivatives}
    \label{fig:fr_grid}
\end{figure}

\begin{figure}[htbp]
    \centering
    % First row of images
    \begin{subfigure}[b]{0.235\textwidth}
        \centering
        \includegraphics[width=\textwidth, trim=18 40 55 55, clip]{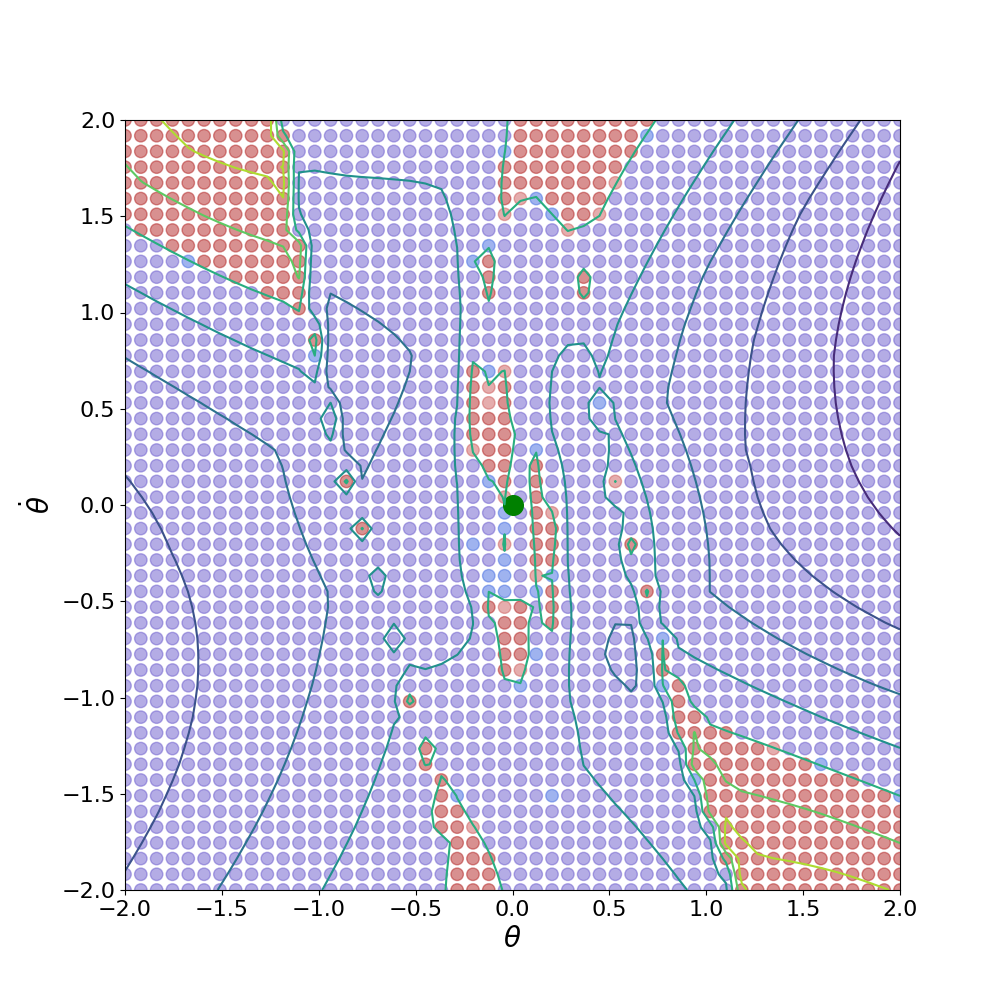}
        \captionsetup{justification=centering}
        \caption{SACLA $(\beta=0.5)$ \\($85.60\% \pm 1.00$)}
        \label{fig:mixed_adv_IP_lvf}
    \end{subfigure}
    \hfill
    \begin{subfigure}[b]{0.235\textwidth}
        \centering
        \includegraphics[width=\textwidth, trim=18 35 55 55, clip]{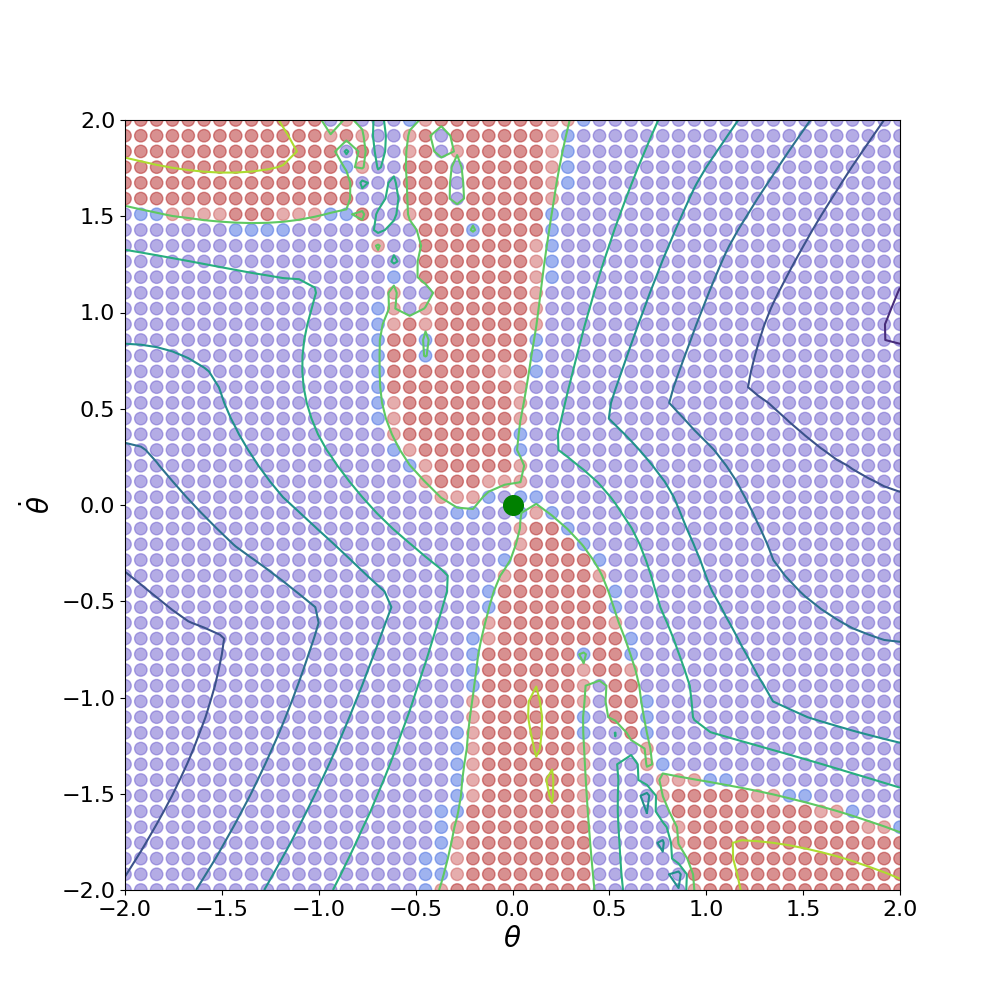}
        \captionsetup{justification=centering}
        \caption{SACLA $(\beta=1)$ \\($76.20\% \pm 0.82$)}
        \label{fig:adv_IP_lvf}
    \end{subfigure}
    
    \hfill
        \begin{subfigure}[b]{0.235\textwidth}
        \centering
        \includegraphics[width=\textwidth, trim=18 35 55 55, clip]{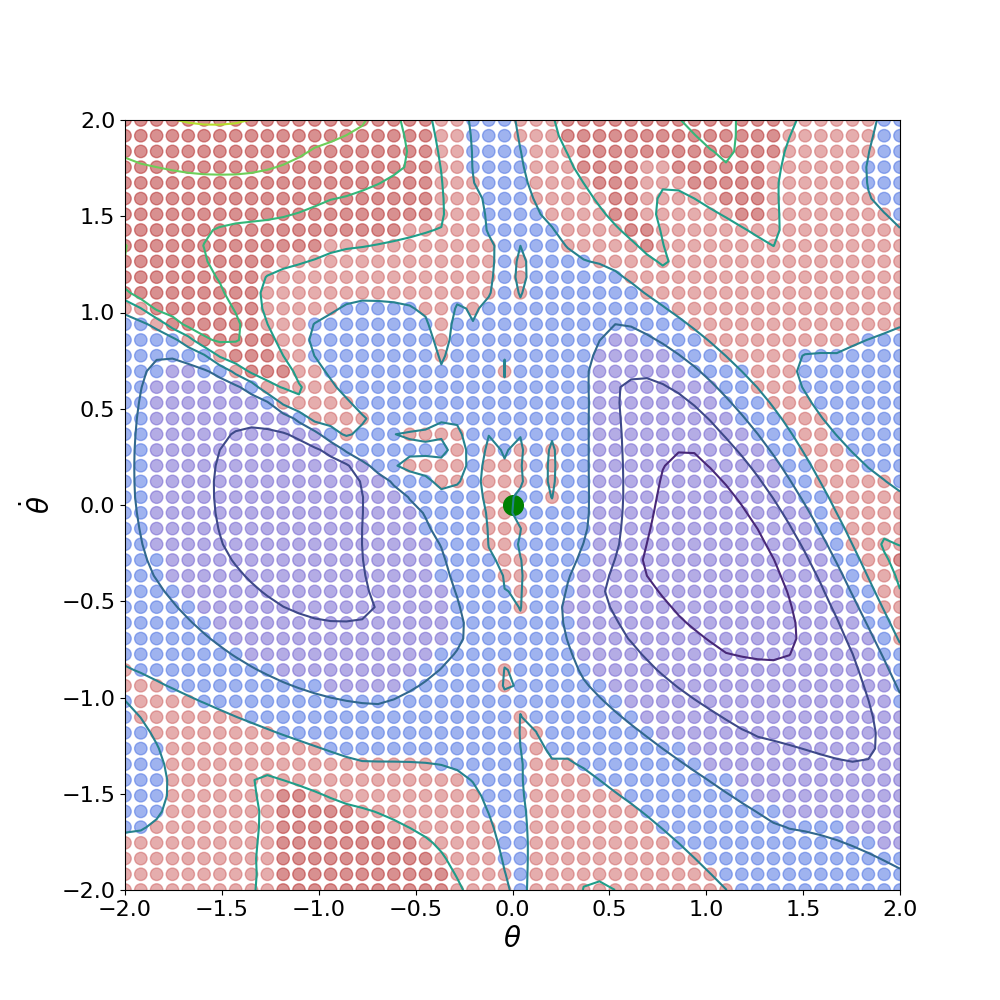}
        \caption{POLYC ($70.84\% \pm 3.40$)}
        \label{fig:polyc_IP_lvf}
    \end{subfigure}
    \hfill
    \begin{subfigure}[b]{0.235\textwidth}
        \centering
        \includegraphics[width=\textwidth, trim=18 40 55 55, clip]{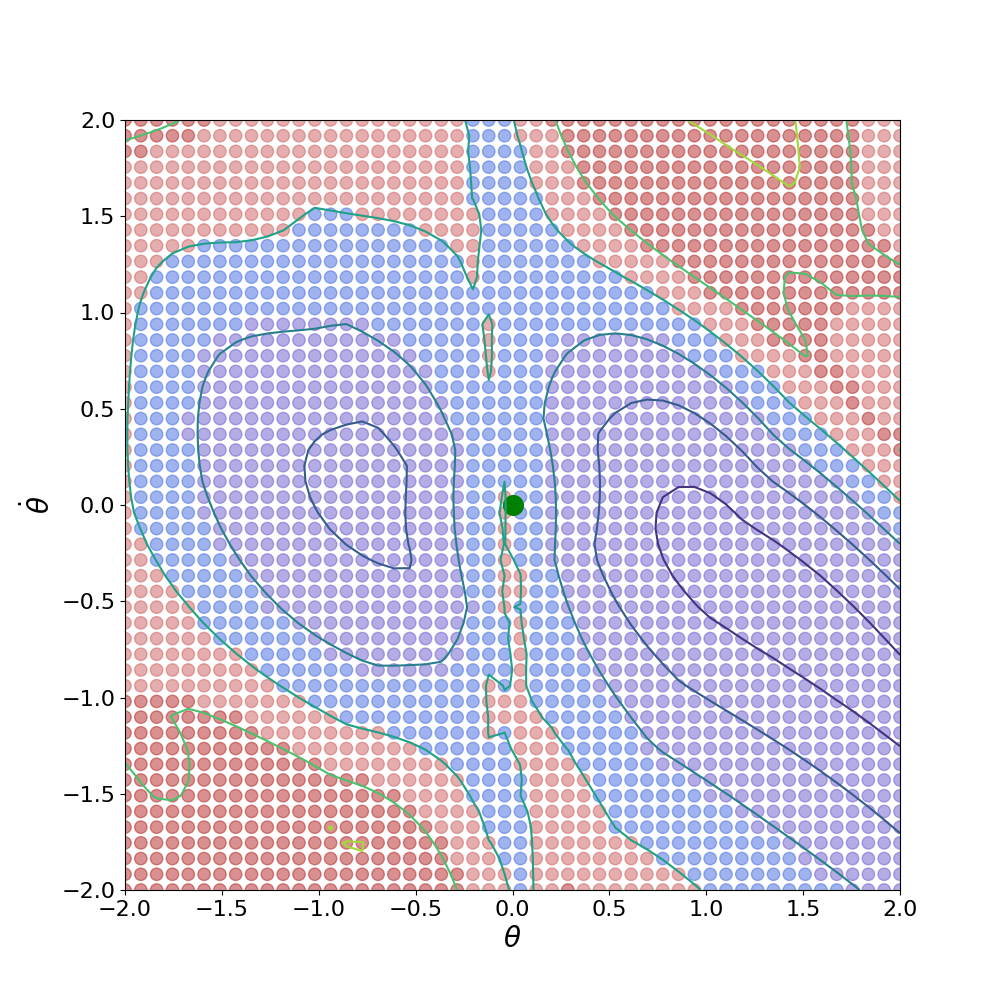}
        \caption{SAC ($60.92\% \pm  3.10$)}
        \label{fig:standard_IP_lvf}

    \end{subfigure}

    \caption{Phase plot of angular displacement ($\theta$) vs angular velocity ($\dot{\theta}$) for SACLA in the \texttt{InvertedPendulum-v4} environment with the final performance percentage and standard deviation. Blue points represent negative Lie derivatives and red points indicate positive Lie derivatives; the intensity of the color indicates the magnitude of the Lie derivative.}
    \label{fig:ip_grid}
\end{figure}

\begin{remark}
The robustness of this probabilistic metric improves with the number of sampled states. As a general rule, the distance between sampled states should be smaller than the maximum potential change in the system's state. The exact number of states needed depends on the system's complexity and the desired confidence level in the stability assessment.
\end{remark}
In both Figure \ref{fig:fr_grid} and Figure \ref{fig:ip_grid} the percentage of negative Lie derivatives was measured using $n = 5000$ points averaged over 5 training seeds. Visualizations use $n = 512$ points for \texttt{FetchReach-v2} \cite{fr_env} and $n = 1000$ points for \texttt{InvertedPendulum-v4} to preserve visual clarity.

%%%% LVF GRID IP
%%%
\subsection{Results and Discussion}
% Intro
We compare SACLA to baseline methods in \texttt{InvertedPendulum-v4} with state space $\mathbb{R}^4$ and the goal-conditioned \texttt{FetchReach-v2} environment with state space $\mathbb{R}^{16}$, focusing on the ROA and Lie derivative distributions to assess controller stability and robustness.
%Fig1
% Figure \ref{fig:fr_grid}

Figure \ref{fig:fr_grid} illustrates the Lie derivative distributions for different methods in the \texttt{FetchReach-v2} environment. Each subplot shows the action vectors and indicates regions with negative (blue arrows) and positive (red arrows) Lie derivatives. SACLA (\(\beta=0.5\)) (88.87\%) achieved the highest ROA, indicating the largest stable region around the goal location. SACLA (\(\beta=1\)) (85.55\%) demonstrates a comparable ROA, although slightly less defined at the goal location due to over-exploration. the SACLA policies both take more curved paths towards the goal, increasing exposure to new states while maintaining stability. Both POLYC (76.75\%) and SAC (79.88\%) are efficient in solving the task, but exhibit limited exploration, resulting in a smaller ROA. The policies tend to have defined regions of similar actions, avoiding new regions of the state space when moving towards the goal, reducing state space coverage.

%Fig 2 ip_grid

Figure \ref{fig:ip_grid} presents Lie derivative distributions for different methods in the \texttt{InvertedPendulum-v4} environment. The phase plots show regions of stability based on the sign and magnitude of the Lie derivatives. SACLA (\(\beta=0.5\)) (85.60\%) achieves the largest percentage of negative Lie derivatives, indicating superior stability across a wide range of states. It is able to strike the ideal balance, achieving a large ROA with well-defined stable regions both near and far from the equilibrium point. SACLA (\(\beta=1\)) (76.2\%) demonstrates advantageous NLF stability in areas far from the equilibrium point but is less defined near the equilibrium. Furthermore, the increased magnitude of Lie derivatives for SACLA can be attributed to the SACLA agent maximizing state-coverage throughout training, which in turn increases the variety in the NLF training data distribution and ultimately the test-time ROA. The strong focus on Lyapunov loss may limit the fine-tuning of policies in the immediate vicinity of the goal. Both SAC (60.92\%) and POLYC (70.84\%) exhibit clearly defined ROAs near the equilibrium point due to repeated exposure during training but show degradation in stability outside this region, we suspect the regularization term in POLYC adds noise during early training, increasing state coverage in this case where the policy learning is faster than the NLF and WM learning. The SACLA methods have reduced final standard deviation, indicating that the NLF exploration term increased the variance in WM training data distribution, leading to more consistent test-time behavior.

\begin{figure}[!htbp]
    \centering
    \begin{subfigure}[t]{0.23\textwidth} % Adjusted width to better use space
        \centering
        \includegraphics[trim=30 22 55 67, clip, scale=0.23]{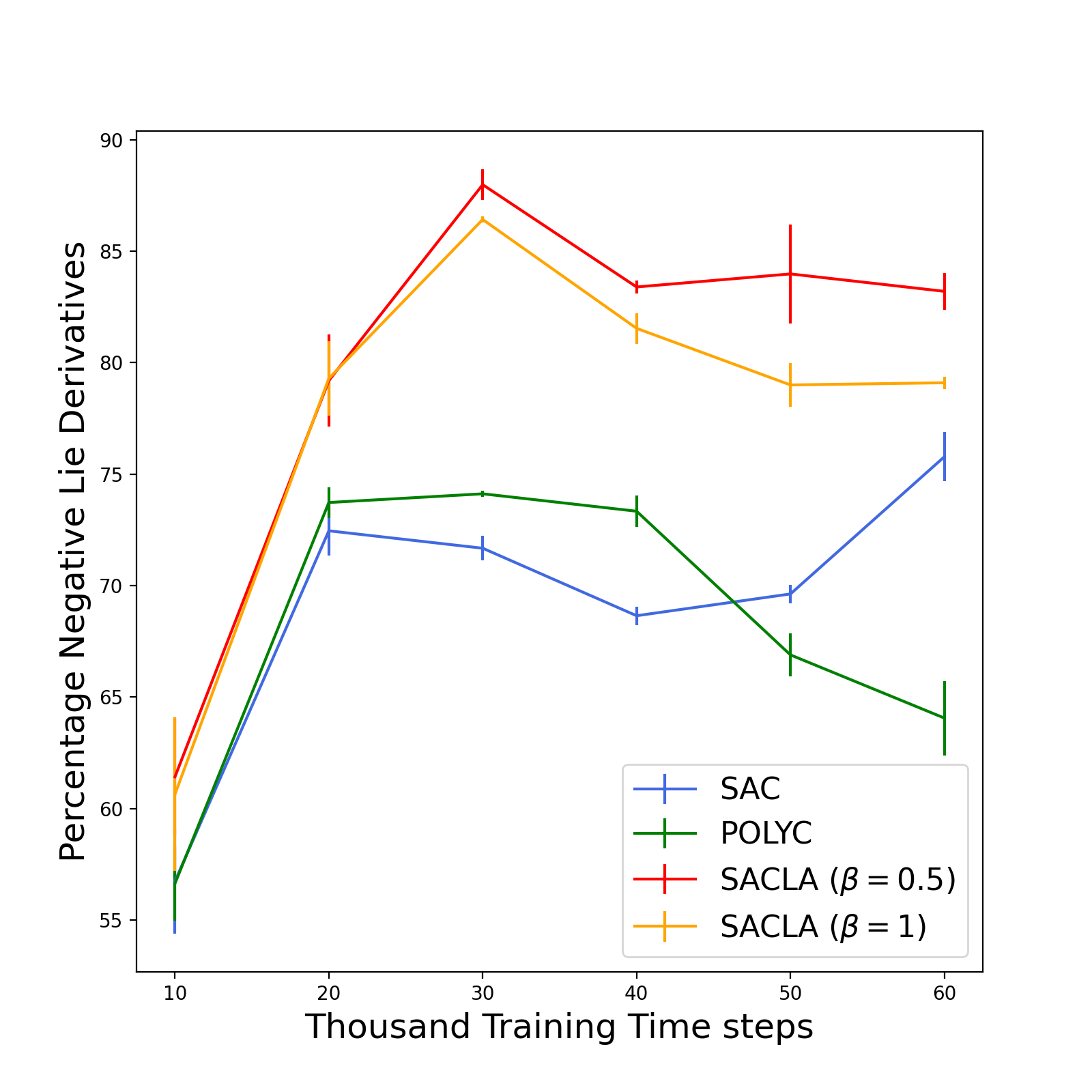}
        \caption{\texttt{FetchReach-v2}}
        \label{fig:trainingReach}
    \end{subfigure}
    \hspace{0.001\textwidth} % Use a controlled space instead of \hfill
    \begin{subfigure}[t]{0.23\textwidth} % Adjusted width to better use space
        \centering
        \includegraphics[trim=25 22 55 68, clip, scale=0.23]{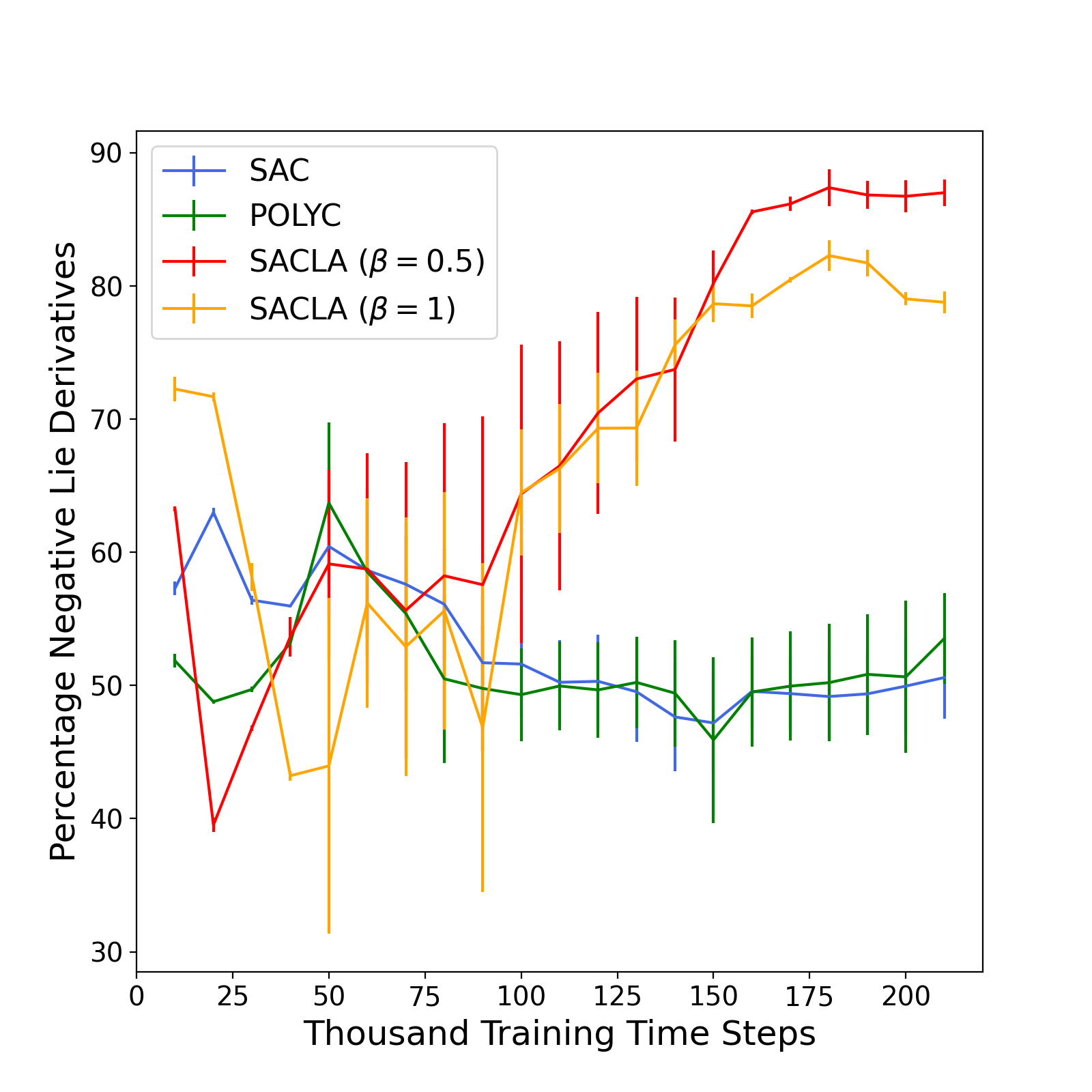} % Make the scale same as the first image
        \caption{\texttt{InvertedPendulum-v4}}
        \label{fig:trainingInv}
    \end{subfigure}
    \caption{Percentage of negative Lie derivatives during the training process for different objective functions, with error bars indicating standard deviation over 15 test seeds}
\end{figure}

% Fig 3 lvf_over_train
% (a)
Figure \ref{fig:trainingReach} presents the percentage of negative Lie derivatives over training time-steps for the \texttt{FetchReach-v2} environment. SACLA (\(\beta=0.5\)) outperforms other methods, achieving the largest ROA with minimal standard deviation across different seeds, as indicated by the error bars representing standard deviation. SACLA (\(\beta=1\)) exhibits a similarly large ROA because it effectively explores the state space while moving towards the goal location. In contrast, SAC, although efficient at solving the task, does not maintain a substantial ROA due to limited exploration. POLYC improves the ROA compared to SAC but still has a smaller ROA than SACLA. The small error bars show in all methods the uncertainty in the WM is relatively low, particularly in SACLA (\(\beta=0.5\)).

% (b)
The percentage of negative Lie derivatives over training time for the \texttt{InvertedPendulum-v4} environment is shown in Figure \ref{fig:trainingInv}. SACLA (\(\beta=1\)) achieves high NLF accuracy in areas far from the equilibrium point, but is less defined near it. SAC and POLYC methods, conversely, show a well-defined ROA near the equilibrium due to repeated exposure but degrade further away. SACLA (\(\beta=0.5\)) strikes the ideal balance, achieving a large ROA with clearly defined regions along with the largest percentage of negative Lie derivatives. The error bars show standard deviation between ROA estimates reduces over time, suggesting that the WM uncertainty, while initially high, reduces until it has a negligible effect of the predicted percentage of negative Lie derivatives. Additionally, the smaller error bars for SACLA objectives indicate that the extra exploration term helps improve the learning of the WM.

% Conclusion
Overall, SACLA (\(\beta=0.5\)), achieves the largest ROA across both environments. This demonstrates that balancing the Lyapunov loss with reward optimization enhances the robustness of learned NLFs, whilst benefiting from the additional guidance of the environment reward.

\subsection{Visualizing Lyapunov Value Evolution} % Move aove Experiments?
\label{sec:stab-analysis}

In addition to the sign of the Lie derivatives, we analyze the distribution of Lyapunov values over time by leveraging the probabilistic outputs of the WM. This provides a visual intuition for how likely the system is to experience unstable states or trajectories are in our system, assisting practitioners in identifying regions of uncertainty in the state space. For notational simplicity, we refer to the refer to $V_\psi$ as $V$. At each time-step we sample $N$ points $\{\tilde{x}_{t+1}^{(i)}\}^N_{i=1}$ from the distribution given by the WM. Then, we calculate the corresponding Lyapunov values $V^{(i)}_t=V(\tilde{x}^{(i)}_{t+1})$ and probability densities $P_t^{(i)}=P(\tilde{x}^{(i)}_{t+1}\mid x_t, u_t)$ using:

\begin{equation}
\label{eq:pdf}
P(\tilde{x}_{t+1} \mid \mu_t, \sigma_t) =\prod^n_{i=1} \frac{1}{\sigma_{t,i} \sqrt{2\pi}} \exp\left(-\frac{(\tilde{x}_{t+1,i}) - \mu_{t,i})^2}{2\sigma_{t,i}^2}\right)
\end{equation}

\begin{figure}[!htbp]
    \centering
    \begin{subfigure}[b]{0.23 \textwidth}
        \centering %lbtr
        \includegraphics[trim=60 40 30 60,clip,width=\textwidth]{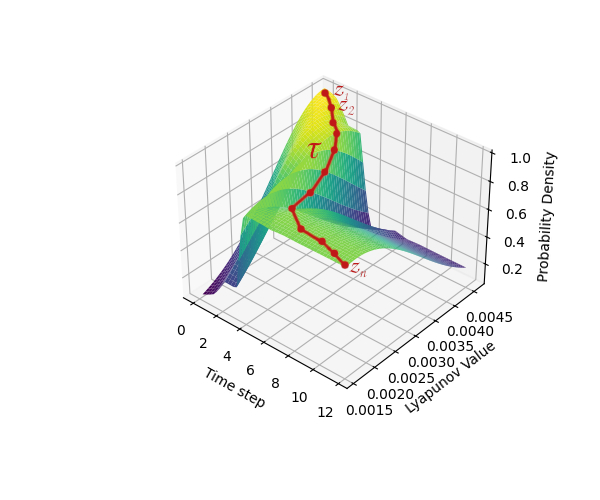}
        \caption{Side view}
        \label{fig:image1}
    \end{subfigure}
    \hfill
    \begin{subfigure}[b]{0.24 \textwidth}
        \centering
        \includegraphics[trim=30 40 30 30,clip,width=\textwidth]{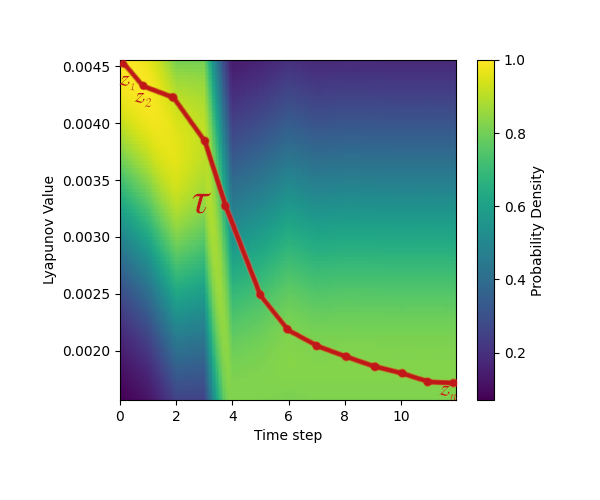}
        \caption{Top View}
        \label{fig:image2}
    \end{subfigure}
    \caption{SACLA ($\beta=0.5$) Lyapunov value distribution over time with an example trajectory on the surface $M$ where $N=100$ for the \texttt{FetchReach-v2} environment.}
    \label{fig:wm_plot}
\end{figure}

To visualize and analyze the trajectories we can model this set of points $(t,V_t^{(i)},P^{i}_t)$ as an empirical approximation of a 2-dimensional surface $M$.
\begin{equation}
    M = \bigcup_{t=0}^{T-1} \left\{ \left( t, V^{(i)}_t, P^{(i)}_t \right) \mid i = 1, 2, \dots, N \right\}
\end{equation}
This surface represents the evolution of Lyapunov values and their associated probabilities over the trajectory. By visualizing $M$, we can analyze the stability properties of the system under the learned policy. For instance, the probability of observing a given point $z_t$ on $M$ under $\pi_\phi$ is given by Equation \eqref{eq:pdf}, whereas the probability of a trajectory $\tau$ given $\tilde{f}_\xi$ and $\pi_\phi$ is:

\begin{equation}
\label{eq:traj-prob}
P(\tau) = p(x_0) \prod_{t=0}^{T-1} P(\tilde{x}_{t+1} \mid x_t, u_t) \pi_\phi(u_t \mid x_t, g_t)
\end{equation}
where $p(x_0)$ is the probability of the initial state. Figure \ref{fig:wm_plot} illustrates how trajectories and states map to curves or points on $M$, and shows how the Lyapunov values change over the state space for a given controller. Regions of low probability indicate the higher WM uncertainty in the state space. Here, the initial conditions are of higher certainty which decrease as the agent moves towards the goal. This is due to the consistency of the manipulators starting position in \texttt{FetchReach-v2}, where as the goal location changes between episodes.

%%%%%%%%%%%%%%%%%%%%%%%%%%%%%%%%%%%%%%%%

\section{Conclusion and Future Work}

We introduced Soft Actor-Critic Lyapunov Approximation (SACLA), a novel method for efficiently learning a Neural Lyapunov Function in highly nonlinear systems. By integrating a World Model with RL and incorporating a self-supervised objective, our approach enables effective exploration for increased test-time stability. Experimental results demonstrate that SACLA ($\beta=0.5$) achieves the largest region of attraction compared to baseline methods, indicating superior stability and robustness. This advancement facilitates the design and verification of stable controllers, particularly in safety-critical applications such as autonomous vehicles and robotic systems.

Despite these promising results, SACLA faces challenges in accumulating prediction errors in Lyapunov value estimations and scalability to larger operational regions. Future research should focus on addressing these limitations by exploring importance sampling techniques to mitigate prediction errors and extending SACLA to encompass broader regions around the goal location. Additionally, integrating methods such as Hindsight Experience Replay \cite{andrychowicz2017hindsight} could enhance sample efficiency in high-dimensional tasks. The learned WM also offers opportunities for constructing control barrier functions and enabling online planning in dynamic environments.

SACLA represents a significant step forward in the efficient learning of accurate Lyapunov functions for complex, nonlinear systems. By bridging the gap between neural approximation methods with traditional stability frameworks, we aim to develop more intelligent and reliable controllers for increasingly complex and safety-critical scenarios.

\section{Acknowledgments}
The first author thanks research funding support from UK Engineering and Physical Science Research Council (project ref: EP/T518050/1) and Veolia Nuclear Solutions

% ------------------------------------------------------------------------------
% \bibliographystyle{unsrtnat}
% \bibliography{references}
\printbibliography[heading=bibintoc]
\end{document}